\documentclass[conference]{IEEEtran}
\IEEEoverridecommandlockouts
\usepackage{cite}
\usepackage{amsmath,amssymb,amsfonts}
\usepackage{algorithm}
\usepackage{graphicx}
\usepackage{algpseudocode}
\usepackage{subcaption}
\usepackage{textcomp}
\usepackage{xcolor}
\def\BibTeX{{\rm B\kern-.05em{\sc i\kern-.025em b}\kern-.08em
    T\kern-.1667em\lower.7ex\hbox{E}\kern-.125emX}}
    
\usepackage{makecell}

\usepackage{booktabs}

\newcolumntype{P}[1]{>{\centering\arraybackslash}p{#1}}
\newcolumntype{M}[1]{>{\centering\arraybackslash}m{#1}}
\begin{document}

\title{On the Adversarial Robustness of Spiking Neural Networks Trained by Local Learning}

\author{Jiaqi Lin, Abhronil Sengupta \\ 
\textit{School of Electrical Engineering and Computer Science} \\
\textit{The Pennsylvania State University} \\ 
University Park, PA 16802, USA \\
Email: \{jkl6467,  sengupta\}@psu.edu
\thanks{}}
\maketitle

\begin{abstract}
Recent research has shown the vulnerability of Spiking Neural Networks (SNNs) under adversarial examples that are nearly indistinguishable from clean data in the context of frame-based and event-based information. The majority of these studies are constrained in generating adversarial examples using Backpropagation Through Time (BPTT), a gradient-based method which lacks biological plausibility. In contrast, local learning methods, which relax many of BPTT's constraints, remain under-explored in the context of adversarial attacks. With the increasing real-world deployment of local learning methods in SNNs, it is critical to investigate whether these approaches introduce adversarial vulnerabilities that could compromise reliability. To address this problem, we examine adversarial robustness in SNNs through the framework of four types of training algorithms. We provide an in-depth analysis showing that direct gradient-based attacks are ineffective for locally trained models. To overcome these limitations, we introduce a hybrid adversarial attack paradigm that leverages the transferability of adversarial instances. The proposed hybrid approach demonstrates superior performance, outperforming existing adversarial attack methods. Furthermore, the generalizability of the method is assessed under multi-step adversarial attacks, adversarial attacks in black-box scenarios, and within the non-spiking domain.
\end{abstract}


\section{Introduction}

\begin{figure*}[htbp] 
\centering
\includegraphics[width=0.99\textwidth]{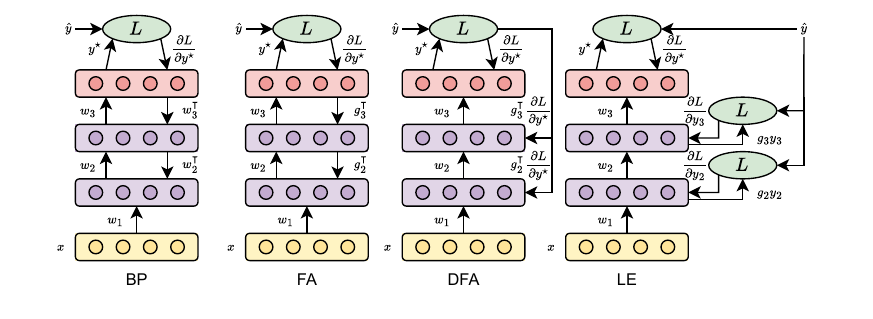}
\caption{A schematic illustrating error propagation pathways in various training methodologies—from global to local learning—includes Backpropagation (BP), Feedback Alignment (FA), Direct Feedback Alignment (DFA), and Local Error (LE). Here, $L$ is the loss function, $x$ is the input instance from the dataset, $y^*$ and $\hat{y}$ are the neural network output and true label respectively, $y_i$ refers to the output of layer $i$, $w_i$ and $w_i^\top$ represent the weight matrix and its transpose respectively for layer $i$, and $g_i$ is the fixed random matrix with the same dimensionality as $w_i$. The incorporation of random matrices to mitigate the constraints of BP-based learning is a key characteristic of FA, DFA, and LE approaches.}
\label{fig:lr_mtd} 
\end{figure*}

Recent advances in neuromorphic computing technologies have addressed the challenges of latency and power consumption inherent in traditional computing chips. Innovations such as Intel's Loihi \cite{davies2018loihi} and the SpiNNaker project \cite{furber2014spinnaker} have made it feasible to deploy Spiking Neural Networks (SNNs) for machine learning applications. SNNs emulate biological neural activity in the brain \cite{maass1997networks, gerstner2002spiking}, where neurons communicate through spike events. This mechanism enables SNNs to function at significantly lower energy levels compared to conventional analog neural networks (ANNs) \cite{diehl2015unsupervised, bi1998synaptic, mead1990neuromorphic, merolla2014million, benjamin2014neurogrid}.

The distinctive spiking dynamics of SNNs have attracted considerable interest in the development of efficient and effective training algorithms, from ANN-to-SNN conversion techniques \cite{sengupta2019going} to backpropagation (BP)-based approaches utilizing the surrogate gradient approximation \cite{neftci2019surrogate}. These works have laid the groundwork for optimizing spiking systems. Although BP-based approaches achieve strong performance, they face criticism for their lack of biological plausibility and computational inefficiency \cite{werbos1990backpropagation, williams1995gradient, decolle_kaiser2020synaptic, eprop_bellec2020solution, lu2022neuroevolution}. These challenges include issues such as the weight transport problem \cite{lillicrap2016random, nokland2016direct, frenkel2021learning}, global error propagation \cite{baldi2017learning, decolle_kaiser2020synaptic}, and reliance on explicit gradient calculations \cite{crick1989recent, lillicrap2020backpropagation}. In response, recent research has investigated alternative local training methods to overcome the limitations of BP-based methods. Figure \ref{fig:lr_mtd} categorizes these local training methods based on their degree of locality. Feedback Alignment (FA) uses fixed random matrices to replace symmetric weight connections during the backward pass \cite{lillicrap2016random}. Direct Feedback Alignment (DFA) relaxes the need for sequential error propagation by transmitting error signals directly from the output layer to each intermediate layer using fixed random matrices  \cite{nokland2016direct, eprop_bellec2020solution}. Local Error Propagation (LE) employs layer-specific error propagation by associating each layer with a cost function, utilizing random matrices to map outputs to pseudo-targets \cite{decolle_kaiser2020synaptic, frenkel2021learning}. Each local learning method addresses concerns related to biological plausibility and relaxes computational requirements, which makes them  potential candidates for training neural networks on-chip.

Advancements in SNN training algorithms have enabled the deployment of SNNs in real-world applications. One significant application area is computer vision, especially in scenarios where temporal information serves as input, such as event cameras \cite{rueckauer2017conversion, mozafari2019bio, paredes2019unsupervised}, LiDAR sensors, or depth cameras \cite{tang2019spiking}, and where computational resources are limited, such as robotics \cite{gutierrez2020neuropod, strohmer2020flexible} and portable devices.
However, to deploy SNNs in real-world situations, noisy, incomplete, and adversarial data pose a critical problem that emphasizes the necessity of the robustness study of SNNs. 
Recent studies have revealed the resilience of SNNs to gradient-based adversarial attacks \cite{sharmin2019comprehensive, liang2021exploring, ding2022snn, kundu2021hire}, using event-based data \cite{buchel2022adversarial, marchisio2021dvs} and various coding mechanisms \cite{nomura2022robustness,lin2022spa, wu2024rsc}. In these explorations, surrogate gradients address the non-differentiability of SNNs in the computation of adversarial gradients \cite{neftci2019surrogate}. 

However, most of them have focused on ANN-to-SNN conversion algorithms \cite{sengupta2019going} or have relied heavily on BP-based approaches \cite{werbos1990backpropagation, rumelhart1986learning}. Few existing works examine the robustness of local learning methods under adversarial attacks. Lin \textit{et al.} \cite{lin2024benchmarking} evaluated SNN robustness with local learning methods, focusing only on linear layers and SNN-specific techniques, which limits generalizability. As SNNs, trained using computationally efficient local learning mechanisms, move closer to deployment in real-world settings \cite{rostami2022eprop}, it is important to assess whether local learning provides inherent robustness or instead introduces new vulnerabilities under adversarial perturbations. 

To address this critical gap, we delve deeper into the adversarial robustness of local learning methods. Our contributions are as follows: 

\begin{enumerate}
    \item First, we demonstrate that in comparison to BP-based approaches, neural networks trained by local learning methods exhibit greater apparent robustness under gradient-based adversarial attacks.
    \item Subsequently, we observe that naive implementations of existing gradient-based attacks, which directly use the adversarial gradients computed by local learning methods as attack directions, are insufficient to produce effective attacks.
    \item To overcome these challenges, we introduce a hybrid attack paradigm, which produces significantly stronger attacks across all benchmark datasets in both spiking and non-spiking domains.
    \item Furthermore, we illustrate the generalizability of the proposed methods to multi-step adversarial attacks, black-box  gradient-based adversarial attacks, and gradient-based adversarial attacks in the non-spiking domain.
\end{enumerate}

\section{Related Works}

\label{sec:convbased}
Presently, two distinct adversarial attack frameworks have been studied for SNNs, namely, conversion-based attack and gradient-based attack. A conversion-based framework involves crafting an adversarial example from an ANN that shares the weights and biases of the corresponding source SNN \cite{fgsm2_sharmin2019comprehensive}. This approach leverages the transferability of adversarial samples between ANNs and SNNs \cite{leontev2021robustness}, enabling successful attacks on SNNs. However, the inefficiency of this method has been highlighted, particularly due to the approximation of spiking dynamics using the ReLU activation function \cite{ding2022snn}. Consequently, our work focuses on gradient-based adversarial attacks for SNNs trained with local learning rules. In gradient-based attacks, the non-differentiable spike function is handled via surrogate gradients during the backward pass with the neuron states unrolled over time \cite{neftci2019surrogate, werbos1990backpropagation, rumelhart1986learning, yin2021accurate}. An alternative gradient-approximation strategy computes gradients based on neuron activities through rate \cite{bu2023rate, ding2022snn, hao2023threaten}, which we also consider in our evaluations.

Sharmin \textit{et al.} \cite{sharmin2019comprehensive} examined conversion-based attacks on both ANN-converted SNNs and SNNs trained from scratch, demonstrating the robustness of SNNs against adversarial attacks and highlighting the robustness advantage of LIF neurons over IF neurons \cite{sharmin2020inherent}. Liang \textit{et al.} \cite{liang2021exploring} explored untargeted gradient attacks, finding that SNNs require greater perturbation for a successful attack. Backdoor trigger attacks have also been a subject of interest. Marchisio \textit{et al.} \cite{marchisio2020spiking} analyzed Spiking Deep Belief Networks under noise attacks, drawing comparisons to backdoor trigger attacks \cite{bd_badnet_gu2019badnets} and reaching similar conclusions. Venceslai \textit{et al.} proposed a fault-injection-based hardware backdoor attack on SNNs \cite{venceslai2020neuroattack}. Jin \textit{et al.} \cite{jin2024data} conducted a comprehensive benchmarking of SNN robustness against backdoor attacks, emphasizing the backdoor migration introduced by conversion-based learning rules. Other studies have examined the unique spiking characteristics of SNNs in adversarial contexts. Time-to-first-spike coded SNNs are shown to be more robust compared to ANNs against adversarial examples \cite{nomura2022robustness}. Similar robustness advantages have also been observed in Poisson coding \cite{nomura2022robustness, lin2022spa} and latency coding \cite{leontev2021robustness}. Furthermore, the robustness of SNNs to event-based adversarial data \cite{buchel2022adversarial} and dynamic vision sensor (DVS) adversarial data \cite{marchisio2021dvs} has been extensively studied, shedding light on their resilience in real-world scenarios.


\section{Preliminaries}
\label{sec:methods}

\subsection{Spiking Neurons}
In this study, spiking neurons follow the dynamics of the Leaky-Integrate-and-Fire (LIF) mechanism \cite{yin2021accurate, neftci2019surrogate, ding2022snn, kundu2021hire}. The membrane potential $u^{t}_i$ with a decay rate of $e^{\frac{-1}{\tau_{\rm mem}}}$ and spike generation $s^{t}_i$ for a layer $i$ at time step $t$ in a discrete time setting are given by:
\begin{equation}
\begin{split}
\label{eq:spiking}
u^{t}_i &= e^{\frac{-1}{\tau_{\rm mem}}} u^{t-1}_i + w_i s^{t}_{i-1} + v_i s^{t-1}_i -  s^{t-1}_i\mathcal{V}_{\rm th} \\
s^{t}_i &= \sigma(u^{t}_i, \mathcal{V}_{\rm th})
\end{split}
\end{equation}
Here, $w_i$ represents the weights of layer $i$. To evaluate the robustness of SNNs against gradient-based adversarial attacks, two types of spiking neurons are utilized in the spiking domain: feedforward spiking neurons (FF), which implement implicit recurrent neuronal activities, and recurrent spiking neurons (REC), which extend the FF model by incorporating weighted spikes from previous time steps to create an explicit recurrent architecture. The expression $v_i s^{t-1}_i$ captures the dynamics of REC neurons at layer $i$, wherein $v_i$ denotes the recurrent weights and $s^{t-1}_i$ signifies the spiking activities at layer $i$ from the preceding time step. In contrast, this term is absent in the dynamics of FF neurons. For both types of neurons, $\sigma(\cdot)$ is the activation function, where a spike is generated when the membrane potential $u^{t}_i$ surpasses the threshold $\mathcal{V}_{\rm th}$. Following each spike generation, the expression $s^{t-1}_i\mathcal{V}_{\rm th}$ captures a subtractive reset mechanism, reducing the membrane potential $u^{t}_i$ by the threshold value $\mathcal{V}_{\rm th}$.

\subsection{Training Methods \label{sec:train_mtd}}

Training a neural network consists of steps to optimize the loss function $L$ with respect to the set of parameters $w$. In this section, we will introduce training algorithms with varying locality to train an SNN. 

\subsubsection{Backpropagation (BP)}

Backpropagation Through Time (BPTT) is a commonly employed BP-based algorithm to optimize SNNs, utilizing their intrinsic recurrent spiking dynamics \cite{werbos1990backpropagation, rumelhart1986learning, neftci2019surrogate, yin2021accurate}. This algorithm works by unrolling neurons across time steps to capture temporal dependencies in neuronal activities, followed by applying backpropagation to the unrolled sequences. The weight update rule for BPTT is defined as follows:
\begin{equation}\label{eq:bp}
\begin{split}
    \Delta w_{ij} &= -\eta \sum_{t=1}^T \frac{\partial L}{\partial s_{i}^t} \frac{\partial  s_{i}^t}{\partial w_{ij}} = -\eta \sum_{t=1}^T  \frac{\partial L}{\partial s_i^{t}} \frac{\partial s_i^{t}}{\partial u_i^{t}} \frac{\partial u_i^{t}}{\partial w_{ij}} \\
    &= -\eta \sum_{t=1}^T  \frac{\partial L}{\partial s_i^{t}} \sigma^\prime(u_i^{t})  \frac{\partial u_i^{t}}{\partial w_{ij}} 
\end{split}
\end{equation}
Here, $\eta$ is the learning rate, $T$ denotes the total number of time steps, and $L$ represents the loss function. $w_{ij}$ corresponds to the weight between two layers $i$ and $j$. At time step $t$, $s_i^{t}$ denotes the spiking events of layer $i$, encoded as binary values where 1 indicates that a spike is fired and 0 indicates that there is no spike activity. The surrogate function approximates the derivative of the non-differentiable spike activation function $\sigma^\prime(u_i^{t})$ with respect to the pre-activation value $u_i^{t}$ at layer $i$.

\subsubsection{Feedback alignment (FA)}
Analogous to the BP-based learning algorithm, FA computes the gradient recursively at each layer. In Equation \ref{eq:bp}, the term $\frac{\partial L}{\partial s_i^{t}}$ captures the recursive propagation of the loss gradient through each layer: 
\begin{equation} 
\frac{\partial L}{\partial s_i^{t}} = \frac{\partial L}{\partial  y^{\star t}} \frac{\partial y^{\star t}}{\partial u_N^{t}} \frac{\partial u_N^{t}}{\partial s_{N-1}^{t}} \frac{\partial s_{N-1}^{t}}{\partial u_{N-1}^{t}} \ldots \frac{\partial s_{i+1}^{t}}{\partial u_{i+1}^{t}} \frac{\partial u_{i+1}^{t}}{\partial s_i^{t}} 
\end{equation} 
where $y^{\star t}$ is the neural network output and $N$ is the total number of layers. Each $\frac{\partial u_{i}^{t}}{\partial s_{i-1}^{t}} \propto w_{i}^\top$. Consequently, gradient calculations in BP-based learning algorithms depend on weights from the forward pass, giving rise to the weight transport problem, which is biologically implausible \cite{nokland2016direct, frenkel2021learning}. To address this issue, FA replaces the weight matrix $w_{i}^\top$ in the backward pass with a fixed random matrix $g_{i}^\top$ \cite{lillicrap2016random}, where $g_{i}^\top$ and $w_{i}^\top$ share the same dimensionality.

\subsubsection{Direct feedback alignment (DFA)}
Relaxing the need for recursive computation of weight gradients, DFA employs a direct backward calculation of gradients from the output layer to each layer \cite{nokland2016direct, eprop_bellec2020solution}. This is achieved by attaching fixed random matrices to map the error signals directly to each layer:
\begin{equation}\label{eq:dfa}
\frac{\partial L}{\partial s_i} = g_{i}^\top \frac{\partial L}{\partial y^\star}
\end{equation}
where $ g_{i}^\top$ is a fixed random matrix mapping the error signals calculated at the output layer $\frac{\partial L}{\partial y^\star}$ to the intermediate layer $i$. The dimensionality of $ g_{i}^\top$ is determined by the shape of the logits at the output layer and the representations of the intermediate layer.

\subsubsection{Local error (LE)}
The LE paradigm extends DFA toward a more localized gradient computation by facilitating layer-wise weight updates \cite{frenkel2021learning, nokland2019training}. This approach enables each layer to learn features based on local information rather than relying on global error signals. An auxiliary cost function evaluates the difference between random readouts $y_i$ and pseudo targets $\hat{y}$ at each layer:
\begin{equation}
    \begin{split}
    \frac{\partial L}{\partial s_i^{t}} &= \frac{\partial L(y_i^t, \hat{y})}{\partial s_i^{t}} \\
    y_i^t &= g_i s_i^t 
\end{split}
\end{equation}
where $g_i$ is a fixed random matrix for layer $i$ of dimensionality determined by the number of neurons in layer $i$ and the shapes of the logits at the output layer.

\subsection{Adversarial Attack Methods}
In this work, we mainly investigate into two types of adversarial attack, namely Fast Gradient Sign Method (FGSM) and Projected Gradient Descent (PGD). FGSM attack leverages the sign of gradients to create adversarial instances that maximize the loss of a neural network, misleading it to make incorrect classifications \cite{fgsm_goodfellow2014explaining}. A data sample $x$ is crafted into adversarial inputs with the perturbation level $\epsilon$ via:
\begin{equation}
\bar{x} = x + \epsilon ~ \textrm{sign}\left(\nabla_x L(f(x,\hat{y}))\right)
\end{equation}
where $L$ is the loss function and $\hat{y}$ is the true label.

PGD is an iterative adversarial attack method designed to create powerful adversarial examples compared to single-step attacks \cite{madry2017towards}. PGD operates by iteratively applying small perturbations to an input, each time moving in the direction of the gradient of the loss function with respect to the input.
\begin{equation}
\bar{x}^{k+1} = \Pi_{\epsilon} \left\{ x^k + \alpha ~ \textrm{sign}\left(\nabla_x L(f(x^k,\hat{y}))\right)\right\}
\end{equation}
Here, $\alpha$ denotes the step size, and the adversarial example is clamped after each step $k$ to ensure that the perturbations are bounded between $-\epsilon$ and $\epsilon$. This iterative process allows PGD to refine the perturbation at each step, making the resulting adversarial examples more effective in attacking the target model.

\subsection{Adversarial Attack Methods for SNNs}
In SNNs, gradient-based adversarial attacks are implemented with surrogate gradients \cite{neftci2019surrogate}, applied to the training methods introduced in Section~\ref{sec:train_mtd}. We consider two strategies: gradients through time (TT) and gradients through rate (TR).

In the TT setting, neuron states are unrolled along the temporal dimension, and the corresponding backward rule is applied across time to obtain the attack direction. We consider four variants: Backpropagation Through Time (BPTT), Feedback Alignment Through Time (FATT), Direct Feedback Alignment Through Time (DFATT), and Local Error Through Time (LETT). For BPTT, the attack direction is obtained through the standard backward pathway, whereas FATT, DFATT, and LETT produce attack directions through their respective local feedback pathways. We refer to the latter quantities as locally computed adversarial gradients.

The TR approach computes gradients directly from the average firing rate of spiking neurons. The firing rate of layer $i$ over a time window $T$ is defined as:
\begin{equation}
    r_i = \frac{\sum_{t=1}^T s^t_{i}}{T}
\end{equation}
where $s_i^t$ denotes the spike at time step $t$. Gradients for the $i$-th layer are then derived through the chain of dependencies between firing rates.
We examine adversarial rate-based variants corresponding to the four training paradigms: Backpropagation Through Rates (BPTR), Feedback Alignment Through Rates (FATR), Direct Feedback Alignment Through Rates (DFATR), and Local Error Through Rates (LETR).

\section{Hybrid Adversarial Attack \label{sec:hyb}}
Although the aforementioned attack methods are effective for BP-based models, our experiments in Section~\ref{sec:influ} show that directly using the locally computed adversarial gradients for adversarial attack generation can be ineffective due to the use of random feedback matrices. The observed gradient obfuscation underscores that evaluating the true robustness of locally trained networks requires appropriate tools. Additionally, in the DFA and LE approaches, locally computed adversarial gradients for a given layer are computed using a layer-specific learning signal, due to the structural characteristics of these methods. To overcome these limitations, we introduce a hybrid method to create adversarial inputs, thereby facilitating more potent attacks against local learning algorithms.
Motivated by the conversion-based adversarial attack (Section \ref{sec:convbased}) \cite{sharmin2019comprehensive}, we propose a hybrid method with the following steps to generate adversarial examples (pseudocode is provided in Appendix~A).
\begin{enumerate}
    \item Initially, given a target model $\textrm{SNN}_{\rm tar}$, a surrogate SNN model $\textrm{SNN}_{\rm sur}$ with global error propagation is initialized with the same topology as $\textrm{SNN}_{\rm tar}$.
    \item Then, the parameter set $w_{\rm tar}$ of the target model is mapped into the parameter set $w_{\rm sur}$ of surrogate SNN.
    \item Subsequently, surrogate gradients are used to address the non-differentiability of spiking function $\sigma(\cdot)$ (Equation \ref{eq:spiking}) in $\textrm{SNN}_{\rm sur}$ during the computation of adversarial gradients $\nabla_x L(f_{w_{\rm sur}}(x,\hat{y}))$.
\end{enumerate}
Fundamentally, our approach constructs the adversarial gradients $\nabla_x L(f_{w_{\rm sur}}(x,\hat{y}))$ using surrogate model following Equation \ref{eq:bp} to replace the locally computed adversarial gradients $\nabla_x L(f_{w_{\rm tar}} (x,\hat{y}))$ computed using localized learning methods. This gradient can be used in any gradient-based attack (e.g., FGSM, PGD), while keeping the remaining update, projection, and clipping steps unchanged. The proposed hybrid attack targets one given local learning model at a time, but it is applicable to multiple local learning paradigms (FA/DFA/LE).

This approach leverages the transferability of adversarial examples \cite{xie2019improving, gu2023survey}, where examples generated for one model can mislead another, even with architectural differences. Although well studied in traditional ANNs, this phenomenon has not been systematically explored in spiking models trained with local learning rules. Our method applies transferability to generate adversarial examples from globally trained models that deceive locally trained SNNs, overcoming the limitations of random feedback paths. Importantly, the surrogate model is used only at inference stage for robustness evaluation and does not affect training, so the biological plausibility and energy-efficiency advantages of local learning are fully preserved. Empirical results confirm the effectiveness of the proposed method.

It is important to distinguish the gradient obfuscation observed in local learning from traditional gradient-masking phenomena. Prior work primarily categorizes gradient masking into shattered, stochastic, and vanishing or exploding gradients \cite{athalye2018obfuscated}. In contrast, the phenomenon studied in our work arises from fixed random feedback matrices. This constitutes a distinct source of gradient obfuscation outside these standard categories. Our hybrid framework is specifically designed to circumvent this mismatch, revealing vulnerabilities that attacks relying directly on the local backward pathway may fail to identify.

\section{Experiments}

\subsection{Dataset and Network Topology \label{sec:setup}}

\begin{table}
\caption{Hyper-parameters for optimal performance on various datasets. LR represents the learning rates and BS represents the batch sizes. DR represents the decay rate of membrane potential $e^{\frac{-1}{\tau_{\rm mem}}}$. \label{tab:param}}
\begin{center}
\begin{tabular}{|l|c|c|c|c|c|c|c|}
\hline
\textbf{Model}& \textbf{Spiking} & DR & $\mathcal{V}_{\rm th}$ & \textbf{LR} & \textbf{BS} & $T$ & \textbf{Epoch}\\
\hline
\multicolumn{8}{|c|}{\textbf{MNIST}} \\
    \hline
     BP FF & Yes & 0.5 & 0.9 & 1e-4 & 8 & 15 & 100\\
     FA FF & Yes & 0.5 & 0.9 & 1e-4 & 4 & 15 & 100\\
     DFA FF & Yes & 0.7 & 0.9 & 1e-4 & 8 & 3 & 100\\
     LE FF & Yes & 0.9 & 0.9 & 1e-3 & 16 & 10 & 100\\
     \hline
     BP REC & Yes & 0.7 & 0.9 & 1e-4 & 8 & 10 & 100\\
     FA REC & Yes &0.7 & 0.9 & 1e-4 & 8 & 10 & 100\\
     DFA REC & Yes & 0.7 & 0.9 & 1e-4 & 8 & 15 & 100\\
     LE REC & Yes & 0.7 & 0.9 & 1e-4 & 8 & 10 & 100 \\
     \hline
     BP ANN & No &  - & - & 1e-4 & 125 & - & 100\\
     FA ANN & No &  - & - & 5e-5 & 125 & - & 100\\
     DFA ANN & No &  - & - & 5e-5 & 125 & - & 100\\
     LE ANN & No &  - & - & 1e-4 & 125 & - & 100\\
    \hline
\multicolumn{8}{|c|}{\textbf{CIFAR-10}} \\
    \hline
    BP FF & Yes & 0.5 & 0.9 & 1e-4 & 250 & 5 & 200\\
     FA FF & Yes & 0.5 & 0.9 & 1e-4 & 128 & 13 & 200\\
     DFA FF & Yes & 0.9 & 0.9 & 1e-4 & 128 & 5 & 200\\
     LE FF & Yes & 0.9 & 0.9 & 7e-5 & 64 & 11 & 200\\
     \hline
     BP REC & Yes & 0.7 & 0.9 & 1e-4 & 250 & 5 & 200\\
     FA REC & Yes &0.5 & 0.9 & 1e-4 & 128 & 13 & 200 \\
     DFA REC & Yes & 0.9 & 0.9 & 1e-4 & 128 & 5 & 200\\
     LE REC & Yes & 0.9 & 0.9 & 7e-5 & 64 & 11 & 200\\
     \hline
     BP ANN & No & - & - & 1e-4 & 150 & - & 100\\
     FA ANN & No &  - & - & 5e-5 & 125 & - & 200\\
     DFA ANN &  No & - & - & 5e-5 & 125 & - & 200\\
     LE ANN & No &  - & - & 1e-4 & 125 & - & 200\\
    \hline
\multicolumn{8}{|c|}{\textbf{DVS Gesture}} \\
    \hline
    BP FF & Yes & 0.6 & 0.9 & 1e-4 & 30 & 30 & 100 \\
     FA FF & Yes  & 0.3 & 0.9 & 1e-4 & 30 & 30 & 200\\
     DFA FF & Yes & 0.6  & 0.9 &  7e-5 & 30 & 30 & 500 \\
     LE FF & Yes & 0.7  & 1.0 &  1e-4 & 10 & 20 & 500 \\
     \hline
     BP REC & Yes & 0.4 & 0.9 & 8e-5 & 30 & 30 & 200\\
     FA REC & Yes & 0.3 & 0.9 & 1e-4 & 30 & 30 & 200 \\
     DFA REC & Yes & 0.6  & 0.9 &  5e-5 & 30 & 30 & 500\\
     LE REC & Yes& 0.9  & 1.0 &  1e-4 & 10 & 30 & 500 \\
    \hline
\end{tabular}
\end{center}
\end{table}

\begin{table}
\centering
     \caption{Accuracy of LeNet-5 across different datasets: (a) spiking LeNet-5 on MNIST and CIFAR-10, (b) non-spiking LeNet-5 on MNIST and CIFAR-10, and (c) spiking convolutional networks on DVS Gesture. Results are reported as mean values over five independent runs, with standard deviations given in parentheses.}\label{tab:perf}
    \subfloat[Spiking LeNet-5 on MNIST and CIFAR-10 datasets]{
        \begin{tabular}{|c|c|c|c|c|}
        \hline
        \textbf{Opt.}&\multicolumn{2}{c|}{\textbf{MNIST}} &\multicolumn{2}{c|}{\textbf{CIFAR-10}}\\
        \cline{2-5} 
        \textbf{Mtd.} & \textbf{\textit{FF}(\%)} & \textbf{\textit{REC}(\%)} & \textbf{\textit{FF}(\%)} & \textbf{\textit{REC}(\%)}\\
        \hline
        BP    & $99.31(0.02)$ & $99.45(0.02)$ & $74.50(0.26)$& $76.59(0.35)$\\
        \hline
        FA   & $99.08(0.03)$& $99.15(0.05)$& $71.16(0.31)$& $73.23(0.13)$ \\
        \hline
        DFA & $98.27(0.07)$& $99.11(0.03)$&  $68.20(0.19)$& $70.09(0.57)$\\ 
        \hline
        LE & $97.94(0.15)$& $98.61(0.10)$&  $63.20(0.37)$& $64.65(0.39)$\\ 
        \hline
        \end{tabular}}
        
    \subfloat[Non-spiking LeNet-5 on MNIST and CIFAR-10 datasets\label{tab:perf_ann}]{
        \begin{tabular}{|c|c |c |}
        \hline
        \textbf{Opt. Mtd.}&\textbf{MNIST} (\%) &\textbf{CIFAR-10} (\%)\\
        \hline
        BP    & $99.30(0.04)$ & $74.41(0.35)$\\
        \hline
        FA   & $99.03(0.03)$& $71.12(0.08)$ \\
        \hline
        DFA &  $98.23(0.08)$& $68.34(0.26)$\\ 
        \hline
        LE & $98.19(0.11)$& $64.61(0.33)$\\ 
        \hline
        \end{tabular}
        }
    
     \subfloat[Spiking Convolutional Networks on DVS Gesture dataset\label{tab:perf_dvs}]{
        \begin{tabular}{|c|c|c|}
        \hline
        \textbf{Opt. Mtd.} & \textbf{\textit{FF}(\%)} & \textbf{\textit{REC}(\%)} \\
        \hline
        BP & 90.00(0.79)  & 90.68(0.74)  \\
        \hline
        FA   & 89.69(0.62) &  90.38(1.43)\\
        \hline
        DFA & 87.05(1.40) & 88.79(1.06) \\ 
        \hline
        LE & 86.29(1.24)& 86.97(1.96)\\ 
        \hline
        \end{tabular}
    }
\end{table}

Our experiments focus on the standard MNIST, CIFAR-10, and DVS Gesture datasets. LeNet-5, consisting of convolutional layers with 64 and 128 channels (each with a kernel size of 3, a stride of 1, and no padding) and two fully connected layers with 120 and 84 neurons \cite{lecun1998gradient}, is evaluated on the MNIST dataset. A wider LeNet-5 with two convolutional layers (64 and 256 channels, kernel size of 3, stride of 1, no padding) and two fully connected layers of 1000 neurons \cite{frenkel2021learning} is tested on the CIFAR-10 dataset. A convolutional network with three convolutional layers (64, 128, and 256 channels; kernel size 3; stride 1; padding 1) and one linear layer mapping features to the output dimension is evaluated on the DVS Gesture dataset.
The architectures are chosen to prevent scaling limitations in local learning methods from affecting comparison fairness \cite{decolle_kaiser2020synaptic, frenkel2021learning}.
The MNIST dataset consists of 60,000 training and 10,000 testing static images of handwritten digits \cite{deng2012mnist}. The CIFAR-10 dataset, containing static images with three channels and a 32-by-32 pixel resolution \cite{cifar10}, presents a more complex classification task compared to MNIST. The DVS Gesture dataset \cite{amir2017low} consists of recordings collected from 29 participants, each performed under three distinct lighting conditions, and is a popular neuromorphic vision dataset. Unlike MNIST and CIFAR-10, which contain static frame-based images, DVS Gesture provides temporally varying event-based data that capture dynamic motion patterns through asynchronous spike events, making it a representative benchmark for evaluating the adversarial robustness of SNNs.

The hyperparameters of the LeNet-5 models trained with various paradigms were optimized for the best performance. Table~\ref{tab:param} summarizes the hyperparameters of optimal performance achieved by each method. The experiments were implemented using the snnTorch and PyTorch libraries in Python and were conducted on an Nvidia RTX 2080 Ti GPU with 11GB of memory. Non-spiking LeNet-5 models are trained to be consistent with the performance of prior works \cite{frenkel2021learning}. The performance of both spiking and non-spiking models on image datasets is presented in Table~\ref{tab:perf}.

Prior to the evaluation of these attack algorithms, the threat model is defined. In the white-box attack scenario, the attacker has complete access to the target neural network's architecture and parameters but lacks access to the original training or testing data. By utilizing model gradients, the attacker generates adversarial examples by applying small, imperceptible perturbations designed to cause incorrect predictions. In the black-box attack scenario, the attacker is assumed to know the target model's architecture but has no access to its parameters or gradients. Instead, adversarial examples are generated using a surrogate model with the same architecture as the target model. For hybrid attack generation, the surrogate model's parameters are mapped to a model that computes attack gradients through global error propagation. This mapping uses only the surrogate model's parameters. The resulting adversarial examples are then tested on the target model.

To evaluate the robustness of neural networks under various attack paradigms, the perturbation level, $\epsilon$, varies from $16/255$ to $96/255$ for the MNIST dataset, from $8/255$ to $16/255$ for the CIFAR-10 dataset, and from $16/255$ to $128/255$ for the DVS Gesture dataset. In the iterative PGD adversarial attack, the number of iterative step is 3, and the step size is $\alpha = 0.07$. The impact of adversarial examples is measured as the loss in accuracy:
\begin{equation}
    {\rm Loss}_{acc} = \frac{{\rm Accuracy}_{\rm clean} - {\rm Accuracy}_{\rm adv} }{{\rm Accuracy}_{\rm clean}}\times 100\%
\end{equation}
where ${\rm Accuracy}_{\rm clean}$ and ${\rm Accuracy}_{\rm adv}$ represent the accuracy of models evaluated on clean samples and adversarial samples, respectively.

\subsection{Influence of Random Matrices in Adversarial Example Generation}

\label{sec:influ}

\begin{figure}
\centerline{\includegraphics[width=0.45\textwidth]{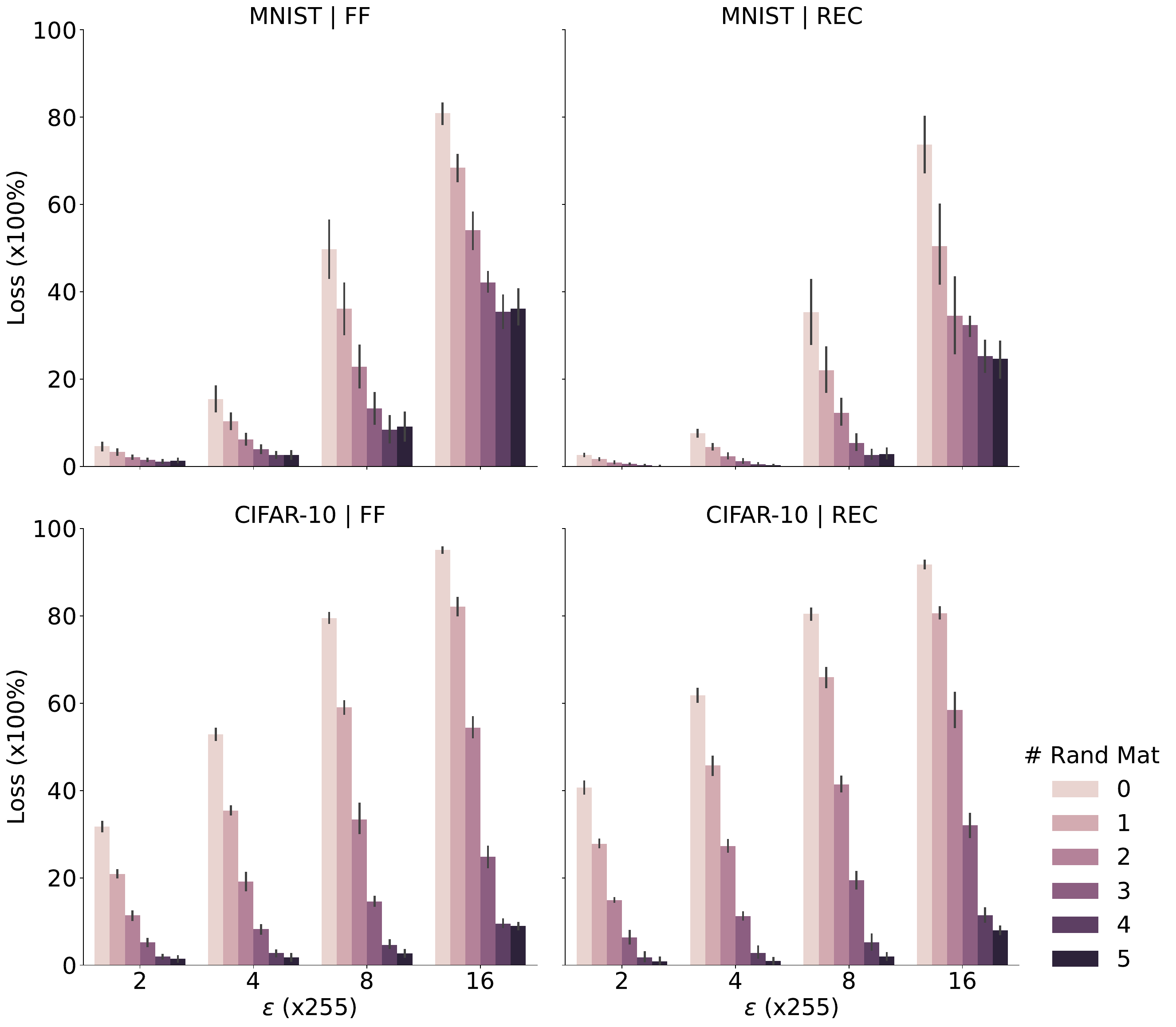}} 
\caption{The influence of random matrices on locally computed adversarial gradients in LeNet-5 models is evaluated under FGSM attacks, using both the MNIST and CIFAR-10 datasets. To progressively increase the use of random feedback, the transposed weight matrices in the backward pass are replaced layer by layer with fixed random matrices, where “\# Rand Mat” denotes the number of such replacements. The loss in accuracy is measured at different perturbation levels. The results are averaged over five independent experimental runs. Increasing replacement of transposed weights with random matrices in backward pass reduces the effectiveness of gradient-based adversarial attacks.}
\label{fig:randmat} 
\end{figure}

In this section, we demonstrate that locally computed adversarial gradients produced using random matrices degrade the effectiveness of gradient-based adversarial attacks. To evaluate this statement, we conducted an experiment comparing BPTT and FATT training methods. \textbf{This experiment is necessary to verify whether standard attacks remain valid under local learning methods, and to motivate the need for an attack paradigm capable of revealing the true vulnerabilities of locally trained networks.} The key difference between BPTT and FATT lies in the backward pass, where BPTT algorithms use transposed weight matrices, and FATT algorithms replace them with random matrices. LeNet-5 is used to assess the impact of random matrices on both the MNIST and CIFAR-10 datasets. As shown in Figure \ref{fig:randmat}, the accuracy of LeNet-5 significantly degrades under the FGSM attack crafted by BPTT as transposed weight matrices are increasingly replaced by random matrices in the backward pass. Notably, the FGSM attack weakens as more random matrices are introduced, saturating at four layers. This observation motivated the development of an attack paradigm, presented in Section~\ref{sec:hyb}, to address the limitations posed by locally computed adversarial gradients. In the next section, we evaluate their practical impact by testing the robustness of SNNs under FGSM attacks.

\subsection{Robustness under FGSM Attack \label{sec:fgsm}}

\begin{figure*}

\centering
\includegraphics[width=0.9\textwidth]{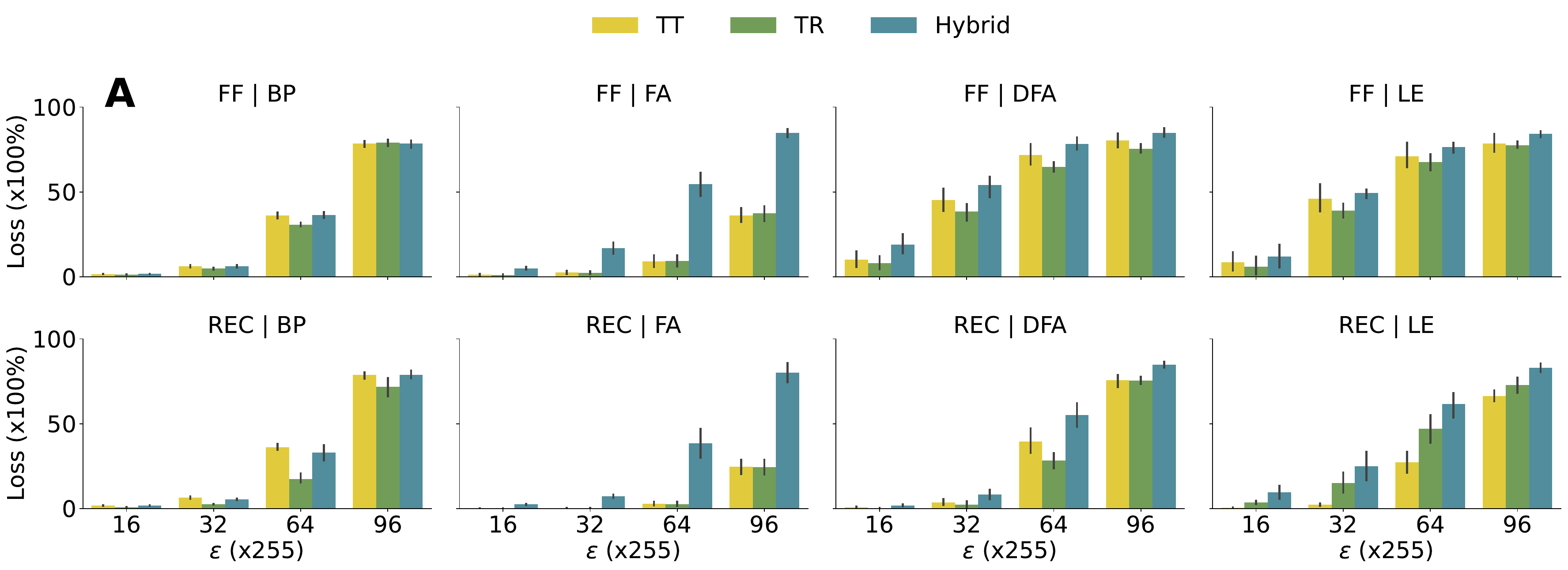}
\hfill
\includegraphics[width=0.9\textwidth]{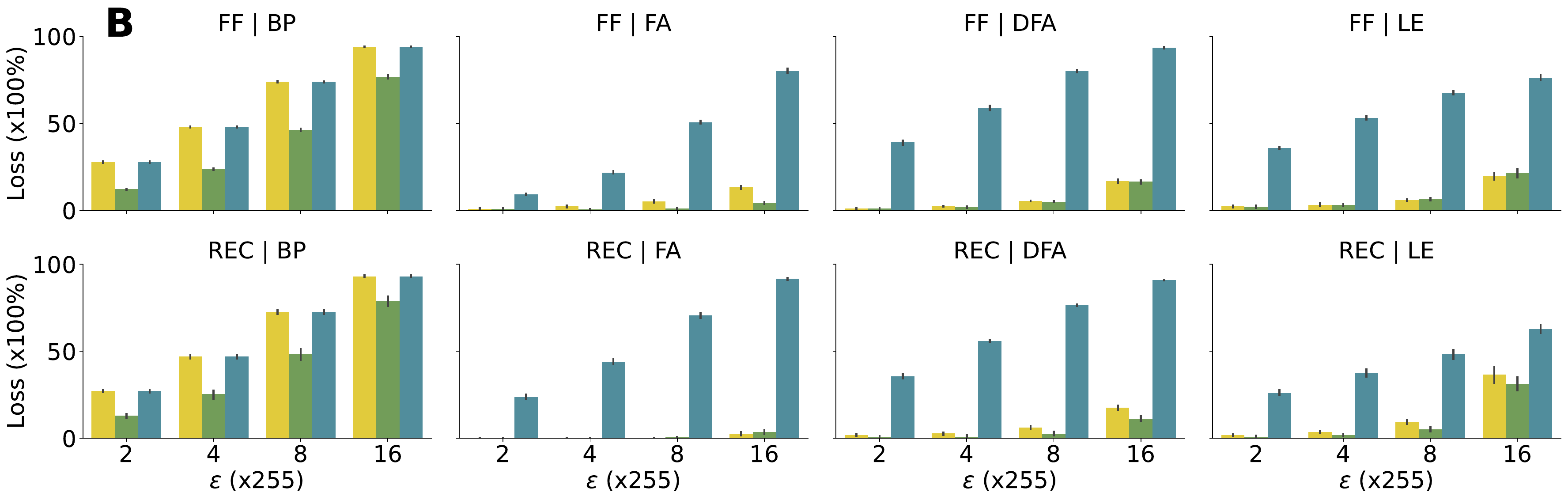}
\hfill
\includegraphics[width=0.9\textwidth]{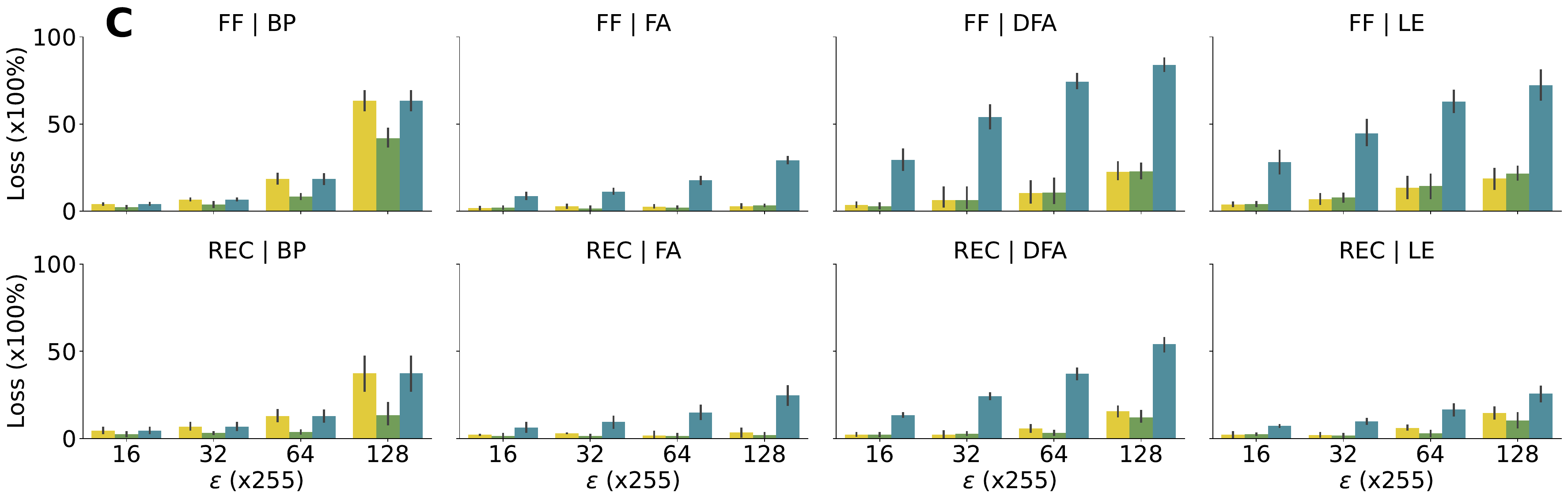}
\caption{ A comparison of the loss in accuracy of SNNs trained by various learning schemes under FGSM attack crafted via TT-based, TR-based, and the proposed hybrid method for three different types of settings: (A) Spiking LeNet-5 on MNIST dataset, (B) Spiking LeNet-5 on CIFAR-10 dataset. (C) Spiking convolutional networks on DVS Gesture dataset. The results are averaged over five independent experimental runs. Local learning methods appear more robust than global ones. Our hybrid FGSM shows the strongest attack among the paradigms. REC SNNs are generally more resilient to hybrid attacks than FF SNNs. }
\label{fig:fgsm}
\end{figure*}

\textbf{Local learning methods initially appear to exhibit greater robustness than global methods (Figure~\ref{fig:fgsm}).} However, this resilience is largely an artifact of gradient obfuscation. Models trained with FA, DFA, and LE demonstrate consistently higher resilience than BP across MNIST, CIFAR-10, and DVS Gesture datasets, with the effect most pronounced on CIFAR-10. These measurements do not reflect true structural robustness.

\textbf{Our proposed hybrid FGSM demonstrates the strongest attack among the three paradigms, functioning as the necessary tool to reveal the true adversarial vulnerabilities of locally trained SNNs.} Here, the gradients are computed via a TT-based approach, as it is more effective when generating adversarial examples using a surrogate SNN model with global error propagation. Figure \ref{fig:fgsm} highlights the superiority of hybrid attacks over TT-based and TR-based FGSM attacks on SNNs. Furthermore, it is observed that REC SNNs exhibit greater resilience to hybrid attacks in comparison to FF SNNs. This trend is consistently observed across the MNIST, CIFAR-10, and DVS Gesture datasets, with the exception of SNNs trained using FA on CIFAR-10. We hypothesize that this observed fluctuation arises from architectural or hyperparameter sensitivity. However, the cause of this exception remains inconclusive and will be investigated in future work.
\subsection{Generalization of Hybrid Attack}

\begin{figure}
\centering
\includegraphics[width=0.45\textwidth]{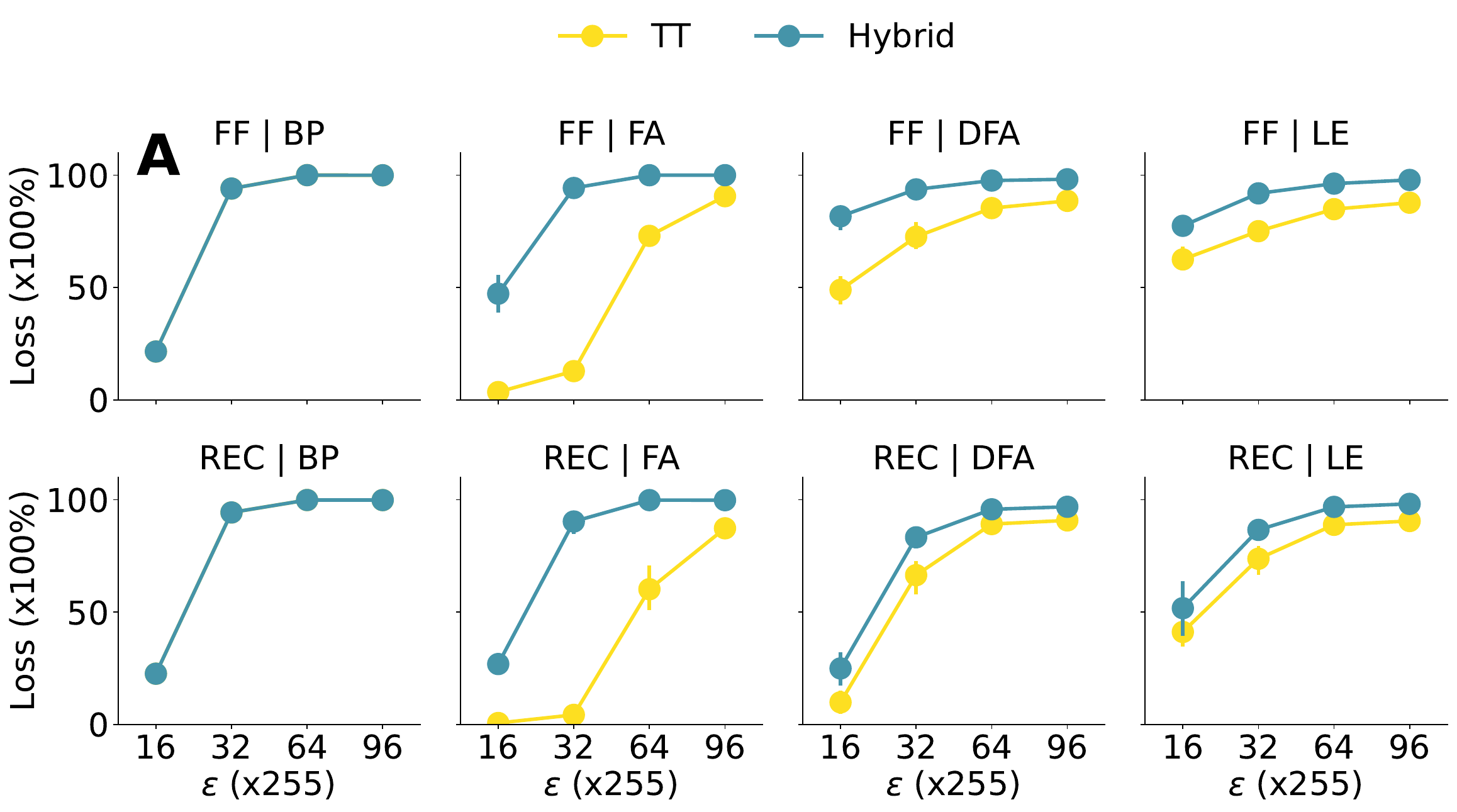}
\hfill
\includegraphics[width=0.45\textwidth]{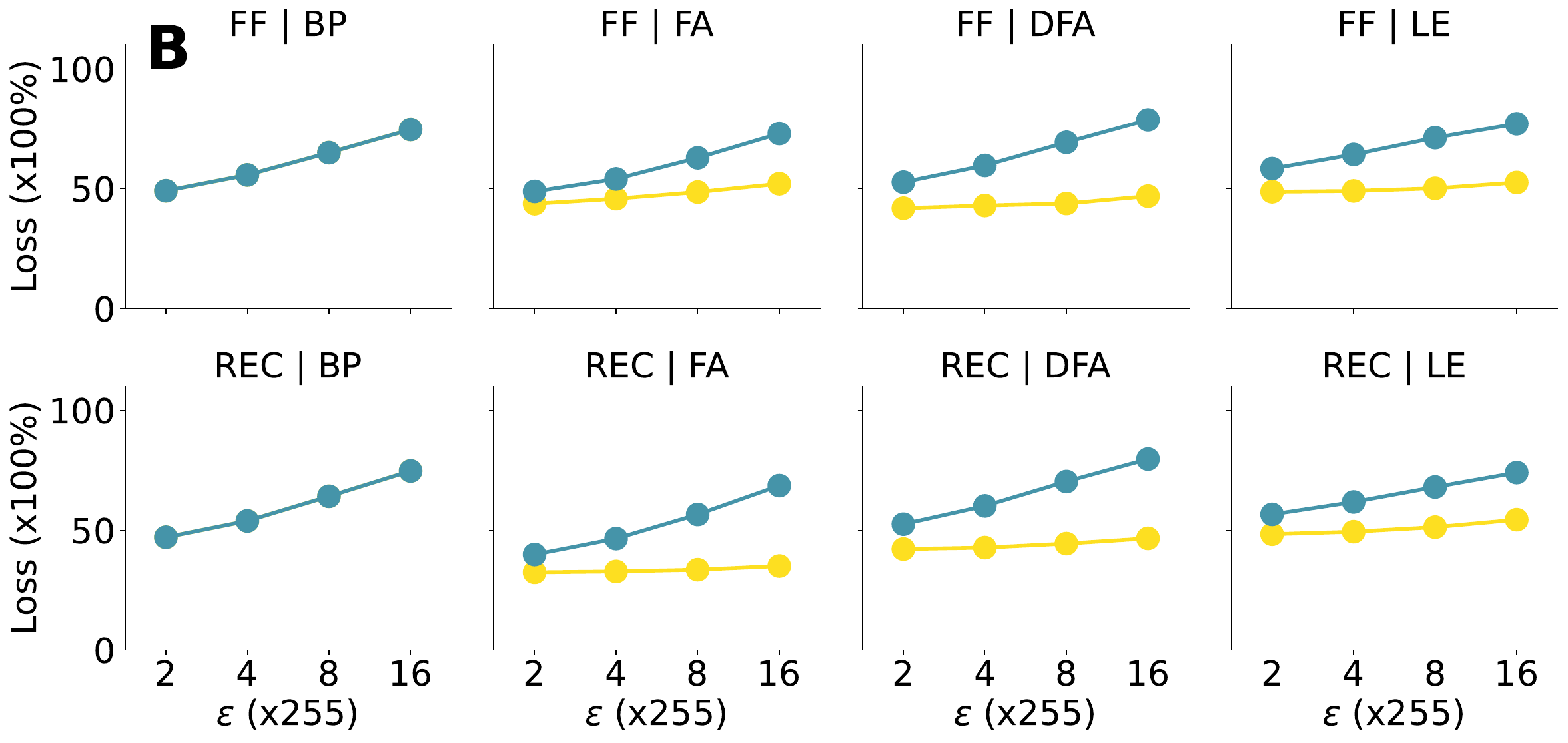}
\caption{ A comparison of the loss in accuracy of SNNs trained using various learning schemes under PGD attacks crafted via TT-based and hybrid methods. This experiment is conducted under two different settings: (A) Spiking LeNet-5 on the MNIST dataset and (B) Spiking LeNet-5 on the CIFAR-10 dataset. The results are averaged over five independent experimental runs.}
\label{fig:pgd}
\end{figure}

\begin{figure}
\centering
\includegraphics[width=0.45\textwidth]{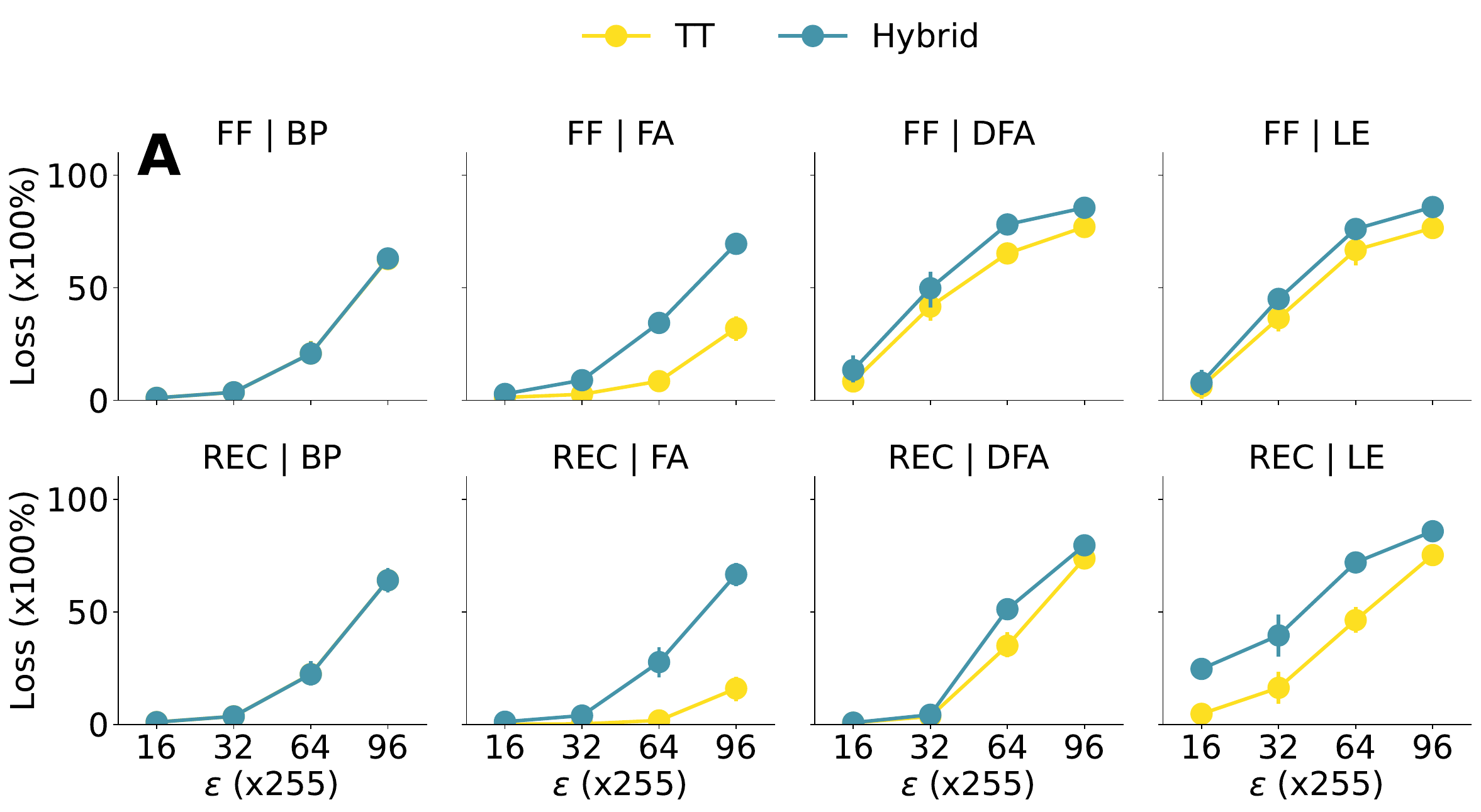}
\hfill
\includegraphics[width=0.45\textwidth]{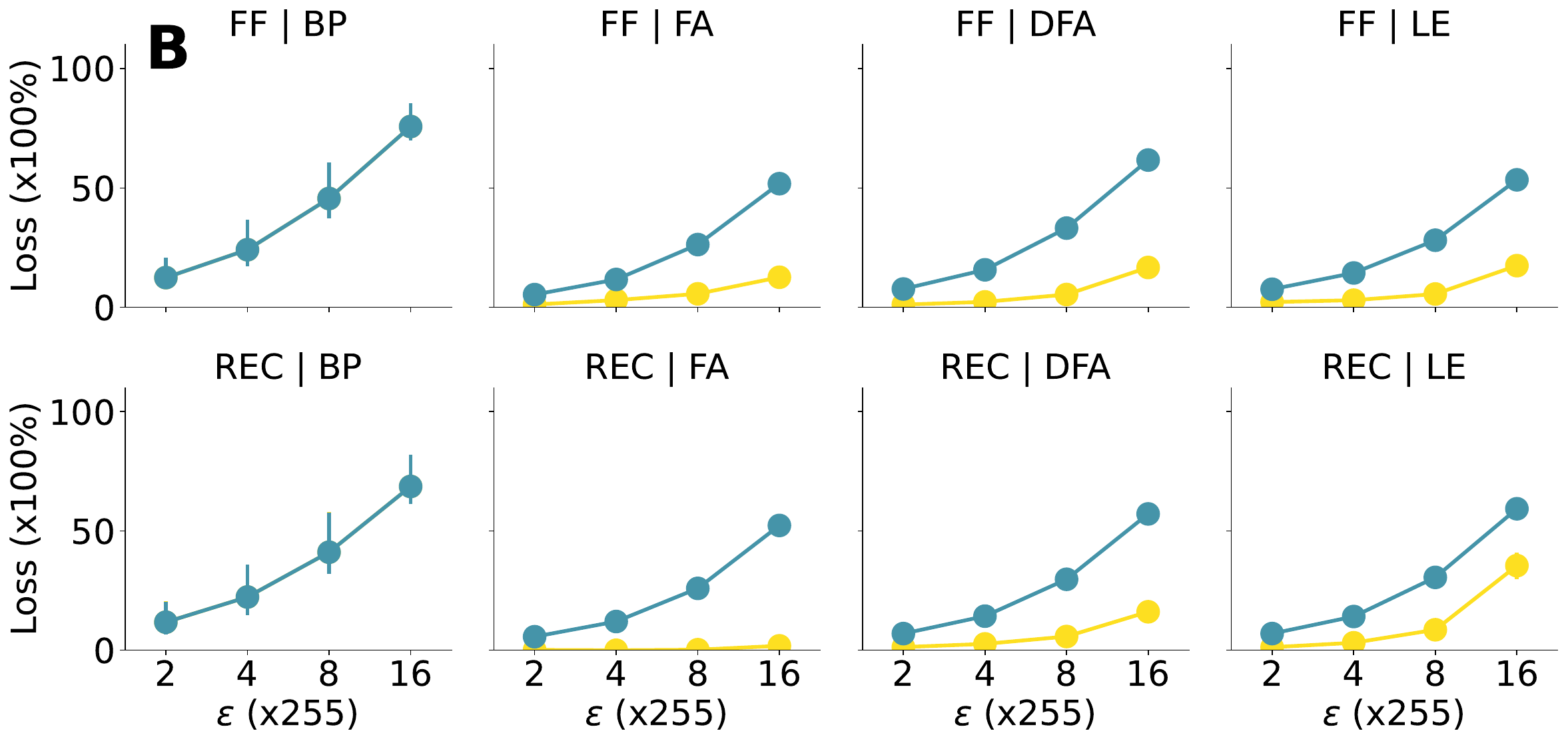}
\caption{A comparison of the loss in accuracy of SNNs trained using various learning schemes under black-box FGSM attacks crafted via TT-based and hybrid methods. This experiment is conducted under two different settings: (A) Spiking LeNet-5 on the MNIST dataset and (B) Spiking LeNet-5 on the CIFAR-10 dataset. The results are averaged over five independent experimental runs.}
\label{fig:fgsmbb}
\end{figure}

\begin{figure}
\centerline{
\includegraphics[width=0.45\textwidth]{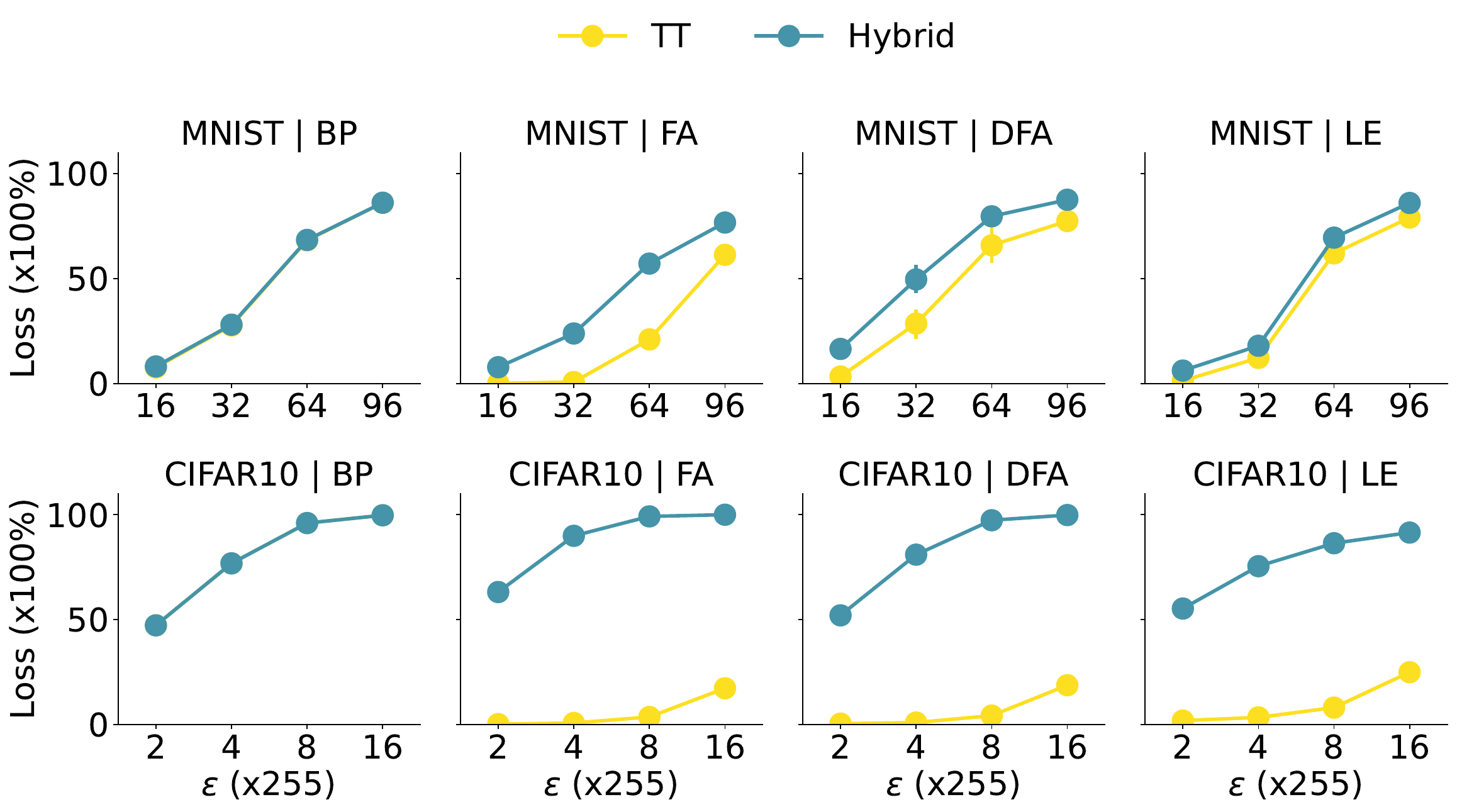}} 
\caption{A comparison of accuracy loss under TT-based and hybrid FGSM attacks in the non-spiking domain is conducted for models trained using various learning schemes. This experiment is conducted under two different settings: (A) Analog LeNet-5 on the MNIST dataset and (B) Analog LeNet-5 on the CIFAR-10 dataset. The results are averaged over five independent experimental runs.} \label{fig:attackann} 
\end{figure}

In the previous section, we established that the proposed attack framework is the necessary tool to provide a more accurate evaluation of SNN robustness under single-step attacks. To rigorously assess these vulnerabilities further, we extend our evaluation across multiple adversarial scenarios.
First, we explore its application to iterative PGD attacks in Figure \ref{fig:pgd}. Gradient-based iterative attacks, such as PGD, have been shown to cause a more significant accuracy degradation compared to their non-iterative counterparts \cite{madry2017towards}. Due to the same rationale in non-iterative adversarial attacks, random matrices alter locally computed adversarial gradients at each PGD step in localized learning methods. 
Next, we demonstrate its performance in a black-box FGSM adversarial attack scenario in Figure \ref{fig:fgsmbb}, where the model parameters are unknown to the attacker. Conventionally, a surrogate model with the same topology as the target model (including the error propagation path) is utilized to create adversarial examples in this case. We map this surrogate model to a globalized surrogate model following the steps in Section \ref{sec:hyb}.
Finally, we assess the hybrid approach in the non-spiking domain in Figure \ref{fig:attackann}. An analog surrogate model is utilized to generate adversarial examples following the steps in Section \ref{sec:hyb}.

Among these scenarios, the same observations can be concluded as in Section \ref{sec:fgsm}. These consistent observations further highlight the necessity of the hybrid framework in addressing the limitations of standard gradient-based attacks. \textbf{Additionally, ANNs are more vulnerable to hybrid-based FGSM attacks compared to SNNs.} The robustness of SNNs can be attributed to the discretization of continuous values in the neuron activation representation, which reduces the influence of adversarial perturbations \cite{sharmin2020inherent}. 

\section{Discussion}

This study examines the robustness of neural networks in both spiking and non-spiking domains using local learning methods. We find that learning algorithms that incorporate random matrices improve robustness against TT-based and TR-based adversarial attacks.
TT-based methods compute gradients across all time steps, while TR-based methods rely on average spiking rates for gradient computation. We empirically demonstrate how random matrices contribute to this apparent robustness and introduce a hybrid adversarial attack paradigm. Empirical results demonstrate the effectiveness of our approach over existing gradient-based approaches.
To further assess the generalizability of the proposed attack paradigm, we evaluate the proposed method across various scenarios, including iterative PGD attacks, black-box gradient-based attacks, and gradient-based adversarial attacks in the non-spiking domain. 
This work provides valuable insights into the broader implications of neuromorphic security, guiding the development of resilient neural networks. Specifically, our findings provide insights into the following areas:
\begin{enumerate}
\item \textbf{Adversarial training.} Standard adversarial training relies on generating adversarial examples via attacks. However, if standard attacks fail to provide informative perturbations due to the presence of random matrices, the model is only optimized against weak adversarial examples, leaving it vulnerable.
\item \textbf{Fair evaluation.} Standard gradient-based attacks are an inadequate benchmark for evaluating locally trained SNNs. This applies equally to claims of adversarial robustness for both new defense mechanisms and novel local learning methods, where researchers should ensure that the reported robustness is not merely an artifact of random feedback matrices.
\item \textbf{Intrinsic Robustness.} While standard gradient-based attacks may overestimate the robustness of FA, DFA, and LE due to locally computed adversarial gradients, other biologically inspired models can exhibit robustness through distinct mechanisms. For example, Equilibrium Propagation attributes robustness to recurrent attractor dynamics in energy-based models \cite{mansingh2024energy}; SNNs benefit from discrete temporal event encoding and nonlinear firing thresholds \cite{sharmin2020inherent}; and predictive coding improves robustness by iteratively denoising inputs toward the clean-data \cite{choksi2021predify}. These studies highlight that robustness in biologically inspired models can arise from multiple mechanisms, motivating careful adversarial evaluation to distinguish genuine robustness from evaluation artifacts.
\end{enumerate}

\section*{Acknowledgments}
This material is based upon work supported in part by the U.S. National Science Foundation under award No. CAREER \#2337646, CCSS \#2333881, CCF \#1955815, and EFRI BRAID \#2318101.

\bibliographystyle{IEEEtran}
\bibliography{iop}
\appendix

\section*{Appendix A. Pseudocode for the Hybrid Attack Method}
\label{app:code}
This appendix provides pseudocode for the proposed hybrid attack method.
Specifically, we introduce a hybrid strategy for crafting adversarial inputs that yields stronger attacks against models trained with local learning. The method can be integrated with any gradient-based attack. Algorithm~\ref{alg:dual} presents the hybrid approach with FGSM as an illustrative example.
\begin{algorithm}
\caption{Hybrid FGSM Attack}\label{alg:dual}
\begin{algorithmic}
\State \textbf{Input:} Clean dataset $X_{\rm clean}$, true label set $Y_{\rm true}$, trained SNN
 model $\textrm{SNN}_{\rm tar}$ with parameter set $w_{\rm tar}$, and adversarial perturbation $\epsilon$
\State \textbf{Output:} Adversarial dataset $X_{\rm adv}$
\end{algorithmic}
\begin{algorithmic}[1]
\State Create a randomly initialized BP-based SNN model $\textrm{SNN}_{\rm sur}$ with the same topology as $\textrm{SNN}_{\rm tar}$ with parameter set $w_{\rm sur}$
\State Set $w_{\rm sur} \leftarrow w_{\rm tar}$ 
\State Create an empty dataset $X_{\rm adv} = \emptyset$
\For {clean sample $x^i \in X_{\rm clean}$ and  and true label $\hat{y}^i\in Y_{\rm true}$}
    \State Output $y^{*,i} = f_{w_{\rm sur}}(x^i)$
    \State Calculate loss $L(y^{*,i}, \hat{y}^i)$
    \State Calculate $\gamma_{\rm adv} \leftarrow \textrm{sign}\left(\nabla_{x^i} L(y^{*,i}, \hat{y}^i)\right)$ 
    \State Adversarial data $\bar{x}^i \leftarrow x^i + \epsilon \gamma_{\rm adv}$
    \State $X_{\rm adv} \leftarrow X_{\rm adv} \cup \{\bar{x}^i\}$
\EndFor
\State Return $X_{\rm adv}$
\end{algorithmic}
\end{algorithm}

\end{document}